\documentclass[10pt]{article} 
\usepackage[preprint]{tmlr}

\usepackage{amsmath,amsfonts,bm}

\def\eqref#1{equation~\ref{#1}}

\def\1{\bm{1}}

\DeclareMathAlphabet{\mathsfit}{\encodingdefault}{\sfdefault}{m}{sl}
\SetMathAlphabet{\mathsfit}{bold}{\encodingdefault}{\sfdefault}{bx}{n}

\newcommand{\Var}{\mathrm{Var}}

\usepackage{hyperref}
\usepackage{url}
\usepackage{graphicx}
\usepackage{enumitem}
\usepackage{amsthm}
\usepackage{amsmath}
\newtheorem{theorem}{Theorem}
\usepackage{tcolorbox}
\usepackage{booktabs}
\usepackage{multirow}
\usepackage{wrapfig}

\definecolor{myblue}{HTML}{6C8EBF}
\definecolor{gblue}{HTML}{7EA6E0}
\definecolor{gred}{HTML}{EA6B66}
\definecolor{in}{HTML}{B0E3E6}
\definecolor{out}{HTML}{B1DDF0}

\title{
Transformation-Augmented GRPO for Enhancing\\Exploration in Reasoning of Large Language Models
}

\author{\name Khiem Le\textsuperscript{1} Phuc Nguyen\textsuperscript{1} Youssef Mroueh\textsuperscript{2} Chi-Heng Lin\textsuperscript{3} Shangqian Gao\textsuperscript{4}\\
        Ting Hua\textsuperscript{1} Nitesh V. Chawla\textsuperscript{1} \\
        \addr \textsuperscript{1}University of Notre Dame, IN, USA \quad \textsuperscript{2}IBM Research \quad \textsuperscript{3}Meta AI \quad \textsuperscript{4}Florida State University, FL, USA
        }

\def\month{MM} 
\def\year{YYYY} 
\def\openreview{\url{https://openreview.net/forum?id=XXXX}} 

\begin{document}

\maketitle

\begin{abstract}
Group Relative Policy Optimization (GRPO) has become the dominant reinforcement learning method for eliciting reasoning capabilities in LLMs using verifiable rewards. Still, it suffers from two critical issues that limit models’ exploration: \textbf{Gradient vanishing} and \textbf{Diversity collapse}. When a training question is too easy or too hard, all sampled responses receive identical rewards, yielding no gradient signal. Meanwhile, the model tends to collapse its responses toward a single reasoning pattern rather than exploring diverse strategies. We propose Transformation-Augmented GRPO (TA-GRPO), a simple yet effective method that addresses both issues via question transformation. For each training question, we automatically create multiple problem-equivalent rephrased questions that alter wording, format, and information order while preserving the underlying intent. As these rephrasings shift the model’s perceived difficulty, responses across the original question and its rephrasings receive mixed rewards and encode more diverse reasoning strategies within it. TA-GRPO jointly computes advantages over this expanded response group and aligns all importance sampling ratios to the original question, enabling the model to learn from a richer set of attempts. Experiments with four LLMs (Qwen3-1.7B/4B-Base and Llama-3.2-1B/3B-Instruct) show that TA-GRPO consistently improves pass@$k$ on competition-level mathematics benchmarks (AMC, OlympiadBench, AIME24, AIME25) and out-of-distribution benchmarks (Minerva, GPQA-Diamond). Notably, it improves the average pass@32 of Qwen3-1.7B-Base and Qwen3-4B-Base by \textbf{4.97}\% and \textbf{4.35}\%, respectively, and matches the pass@32 of baselines trained with up to 2.5× more data.
\end{abstract}

\section{Introduction}
The rise of reasoning-centric Large Language Models (LLMs), such as OpenAI-o1 \citep{jaech2024openai} or DeepSeek-R1 \citep{guo2025deepseek}, has led to significant success in solving complex questions in mathematics and coding. The driver is reinforcement learning with verifiable rewards (RLVR) \citep{lambert2024tulu}, which trains the model by maximizing automatically scored rewards such as the correctness of final answers. Under RLVR, GRPO \citep{shao2024deepseekmath} has emerged as the most widely adopted policy optimization method: for each training question, it samples $G$ responses, computes their advantages via group-relative reward normalization, and updates the model with the corresponding importance sampling ratios.

Despite rapid progress, most GRPO advances \citep{liu2026advances, surana2026} overlook two issues that constrain exploration in the model's reasoning: \textbf{Gradient vanishing} and \textbf{Diversity collapse}. Gradient vanishing occurs when a training question is too easy or too hard for the model: all $G$ sampled responses receive identical rewards, diminishing advantages and yielding no gradient signal. An exception, DAPO \citep{yu2026dapo}, applies dynamic sampling to replace such questions, but trades away a significant portion of training data. Diversity collapse \citep{yun2025price} arises when the model collapses to a single reasoning pattern across $G$ responses rather than exploring diverse strategies. Classical entropy regularization techniques (including one in DAPO) cannot control it because of the vast action space of an LLM \citep{jiang2025rethinking}.

\begin{figure}[!t]\centering\includegraphics[width=\linewidth]{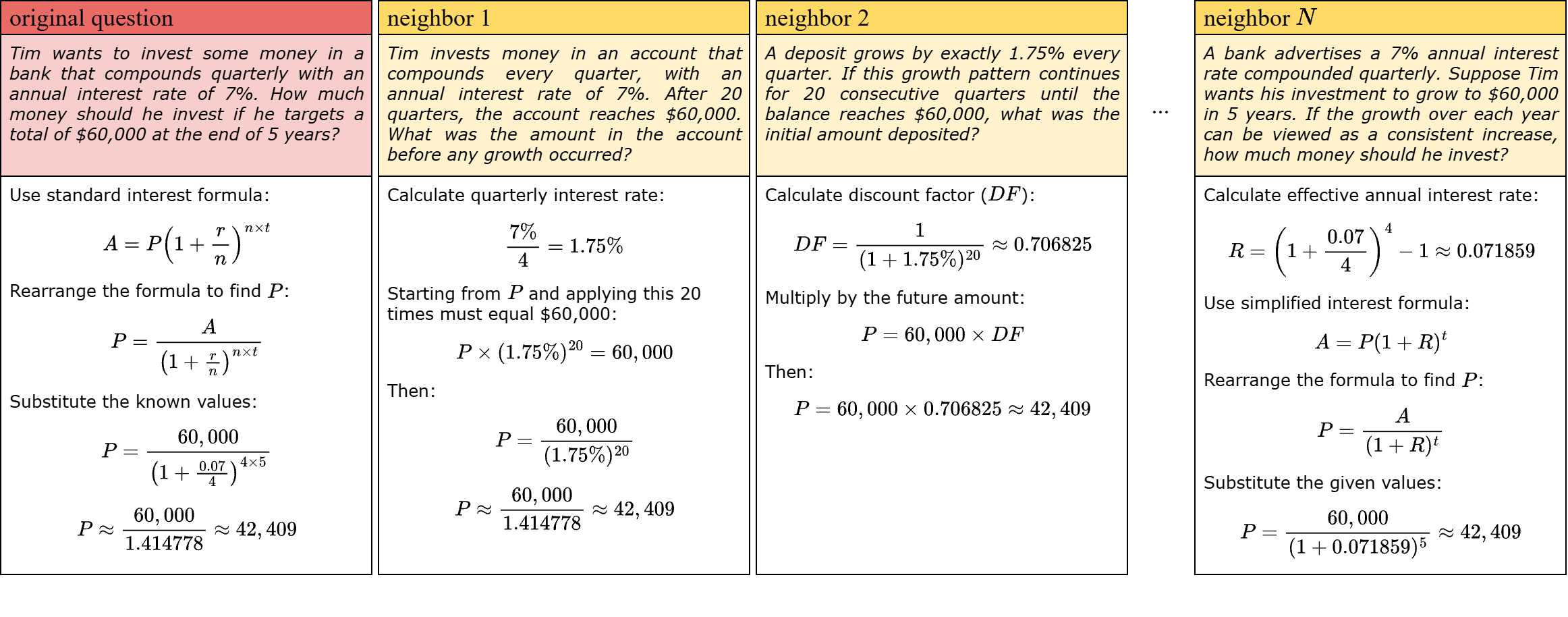}
\caption{An example showing how different rephrasings of a question induce diverse reasoning strategies. Each column contains a question and the model’s response. The model adopts a distinct strategy for each rephrasing: standard interest formula (original), calculate quarterly interest rate (neighbor 1), calculate discount factor (neighbor 2), and calculate effective annual interest rate (neighbor $N$). Despite these differences, all attempts arrive at the same final answer (\$42,409), since the rephrasings preserve the underlying intent.}\label{fig-1}
\end{figure}

In this work, we propose Transformation-Augmented GRPO (TA-GRPO), a simple yet effective policy optimization method that addresses both issues through question transformation. Prior studies found that linguistic perturbations, such as changes in wording, format, or information order \citep{salinas2024butterfly, chatterjee2024posix, guan2025order}, can substantially alter the model's interpretation of a question. Motivated by this finding, we transform each question in the training data $N$ times by rephrasing it from different perspectives while preserving the underlying intent. This creates $N$ problem-equivalent rephrased questions, serving as $N$ “neighbors” of the original question. As shown in Figure \ref{fig-1}, different rephrasings lead the model to adopt distinct solution strategies.

Building on these insights, TA-GRPO modifies the standard GRPO pipeline as follows. For each training question, TA-GRPO samples from the model $G$ responses to the original question and $G$ responses to each of its $N$ rephrasings, forming an expanded group of $(N + 1) \times G$ responses. Then, TA-GRPO jointly computes advantages over this expanded group via \textbf{joint} group-relative reward normalization. To anchor the update to the original question, all importance sampling ratios are conditioned on the original question regardless of which rephrasing produced each response. This design directly tackles both issues. By varying the model's perceived difficulty, the rephrasings increase the likelihood that the expanded group has responses that receive mixed rewards, even when the original question alone would not, thereby mitigating gradient vanishing, and encode more diverse reasoning strategies within it, thereby mitigating diversity collapse.

Our contributions are summarized as follows:
\begin{itemize}[topsep=0em, leftmargin=*, noitemsep]
\item \textbf{Targeting two overlooked issues in GRPO at once.} Prior GRPO advances either omit addressing gradient vanishing and diversity collapse or treat them in isolation rather than in tandem.
\item \textbf{TA-GRPO.} Our simple yet effective GRPO extension concurrently addresses two issues via question transformation, pooling responses across each training question and its $N$ rephrasings into an expanded group, jointly computing advantages and anchoring importance sampling ratios to the original question.
\item \textbf{Evaluation.} TA-GRPO consistently improves pass@$k$ across four LLMs. On competition-level mathematics benchmarks (AMC, OlympiadBench, AIME24, AIME25), average pass@32 gains reach \textbf{5.09}\%/\textbf{4.76}\% for Qwen3-1.7B/4B-Base and \textbf{3.50}\%/\textbf{5.25}\% for Llama-3.2-1B/3B-Instruct; on out-of-distribution benchmarks (Minerva, GPQA-Diamond), gains reach \textbf{4.74}\%/\textbf{3.53}\% and \textbf{3.40}\%/\textbf{2.68}\%.
\item \textbf{In-depth analysis.} We showcase that TA-GRPO can serve as an economical alternative to collecting additional high-quality training data, empowering models to achieve pass@32 comparable to that obtained by baselines with up to 2.5× more data.
\end{itemize}

\section{Related Work}
Since its introduction, research on GRPO \citep{liu2026advances, surana2026} has proceeded along two complementary directions, which we review in turn: \textbf{Understanding} that characterizes its practical and theoretical properties, and \textbf{Advances} that propose algorithmic modifications.

\textbf{GRPO Understanding} - A line of work seeks to characterize GRPO practically and theoretically. \citet{mroueh2025reinforcement} shows that, with a slight calibration, the GRPO objective has an adaptive and weighted contrastive form. Meanwhile, \citet{wu2025takes} establish a fundamental connection between the GRPO objective and the direct preference learning objective \citep{rafailov2023direct}. \citet{zhou2026demystifying} present a unified framework based on classical U-statistics, characterizing the mean squared error of the GRPO gradient, deriving finite-sample error bounds, and showing the asymptotic distribution of the suboptimality gap. \citet{yao2026group} further demystify the role of clipping in the GRPO objective, revealing that enlarging the clipping range far beyond conventional choices is often viable and can accelerate convergence. These analyses mainly focus on properties of the existing GRPO objective; our contribution is orthogonal, making algorithmic modifications, specifically at the question- and response- levels.

\textbf{GRPO Advances} - A parallel line of work aims to make GRPO better. Given $G$ sampled responses per training question, SA-GRPO \citep{han2026self} leverages the model’s confidence to complement the response-level rewards, while SAGE-GRPO \citep{huang2026does} seeks to control the excessive lengths of these responses. Dr. GRPO \citep{liu2025understanding} aims to overcome the question-level difficulty bias in GRPO, whereas GRPO-LEAD \citep{zhang2025grpo} embraces this bias by reweighting each question by its difficulty, estimated via the inverse success rate of the $G$ responses. SEED-GRPO \citep{chen2025seed} strengthens the model’s comprehension of the question by amplifying semantic consistency among the $G$ responses, and ExGRPO \citep{zhan2026exgrpo} integrates an efficient experience replay mechanism to bootstrap the training. Despite this rapid progress, two issues remain unresolved. Gradient vanishing: DAPO \citep{yu2026dapo} circumvents it via dynamic sampling, but at the cost of discarding too-easy and too-hard questions, which constitute a significant portion of training data. Diversity collapse: Classical entropy regularization techniques \citep{ziebart2010modeling, haarnoja2018soft}, including one employed in DAPO, struggle to govern it, owing to the vast action space induced by an LLM's vocabulary, which often exceeds 100,000 tokens \citep{jiang2025rethinking}. Our work targets both issues directly, without relying on these ineffective approaches.

\section{Revisiting GRPO}
Recent efforts to strengthen the reasoning capabilities of LLMs have increasingly relied on GRPO, which dispenses with a critic model required by predecessors \citep{schulman2015trust, schulman2017proximal} and derives advantages from group-relative reward normalization. Let $\pi_\theta$ represent the model, which is parameterized by $\theta \in \mathbb{R}^d$. Given the question $q$ drawn from training data $\mathcal{D}$, the model generates $G$ responses to it as a group $\{o_i \sim \pi_\theta (\cdot|q)\}_{i=1, ..., G}$. Denote their rewards as $\{R_i\}_{i=1, ..., G}$, GRPO computes the advantages of $G$ responses via:
\begin{equation}
\Big\{A_i = \frac{R_i - \text{mean}(\{R_i\}_{i=1, ..., G})}{\text{std}(\{R_i\}_{i=1, ..., G})+\varepsilon}\Big\}_{i=1, ..., G}. 
\label{eqn-1}
\end{equation}
Following that, GRPO updates the model by minimizing an objective defined over these advantages and the corresponding importance sampling ratios:
\begin{align}
&\mathfrak{J}(\theta)=-\mathbb{E}_{q \sim \mathcal{D}}[J(\theta; q)], \nonumber \\
&J(\theta; q) = \left[ \frac{1}{G} \sum_{i=1, ..., G} \frac{1}{|o_i|} \sum_{t=1, ..., |o_i|} \Bigg( \text{min}\Big( r_\theta(o_{i, t})A_i, \text{clip}\big( r_\theta(o_{i, t}), 1-\epsilon, 1+\epsilon \big)A_i \Big) \Bigg) \right], \nonumber \\
&\text{where} \quad r_\theta(o_{i, t}) = \frac{\pi_\theta(o_{i, t} | q, o_{i, < t})}{[\pi_\theta(o_{i, t} | q, o_{i, < t})]_{\text{not\_updated}}}. 
\label{eqn-2}
\end{align}
Intuitively, GRPO enables the model to learn from $G$ sampled responses without a critic model, instead estimating the baseline from the rewards.

Despite its strengths, this group-relative reward normalization design also introduces two structural vulnerabilities that bottleneck exploration in reasoning of the model \citep{yue2025does, akgul2026rethinking, shojaee2026illusion}: \textbf{Gradient vanishing} and \textbf{Diversity collapse}. Gradient vanishing: When the question $q$ is too easy or too hard (all $G$ sampled responses succeed or fail, respectively), the reward satisfies $R_i = \text{mean}(\{R_i\}_{i=1, ..., G})$ for all $i$, leading to $A_i = 0$ for all $i$, yielding $\nabla_{\theta}J(\theta; q)=0$. Diversity collapse: Since $\{o_i\}_{i=1, ..., G}$ are sampled solely from $\pi_\theta (\cdot|q)$; once $\pi_\theta (\cdot|q)$ concentrates around a confident but narrow mode, $G$ responses to $q$ are progressively collapsed.

\section{Transformation-Augmented GRPO}
\begin{figure}[!t]
\centering\includegraphics[width=\linewidth]{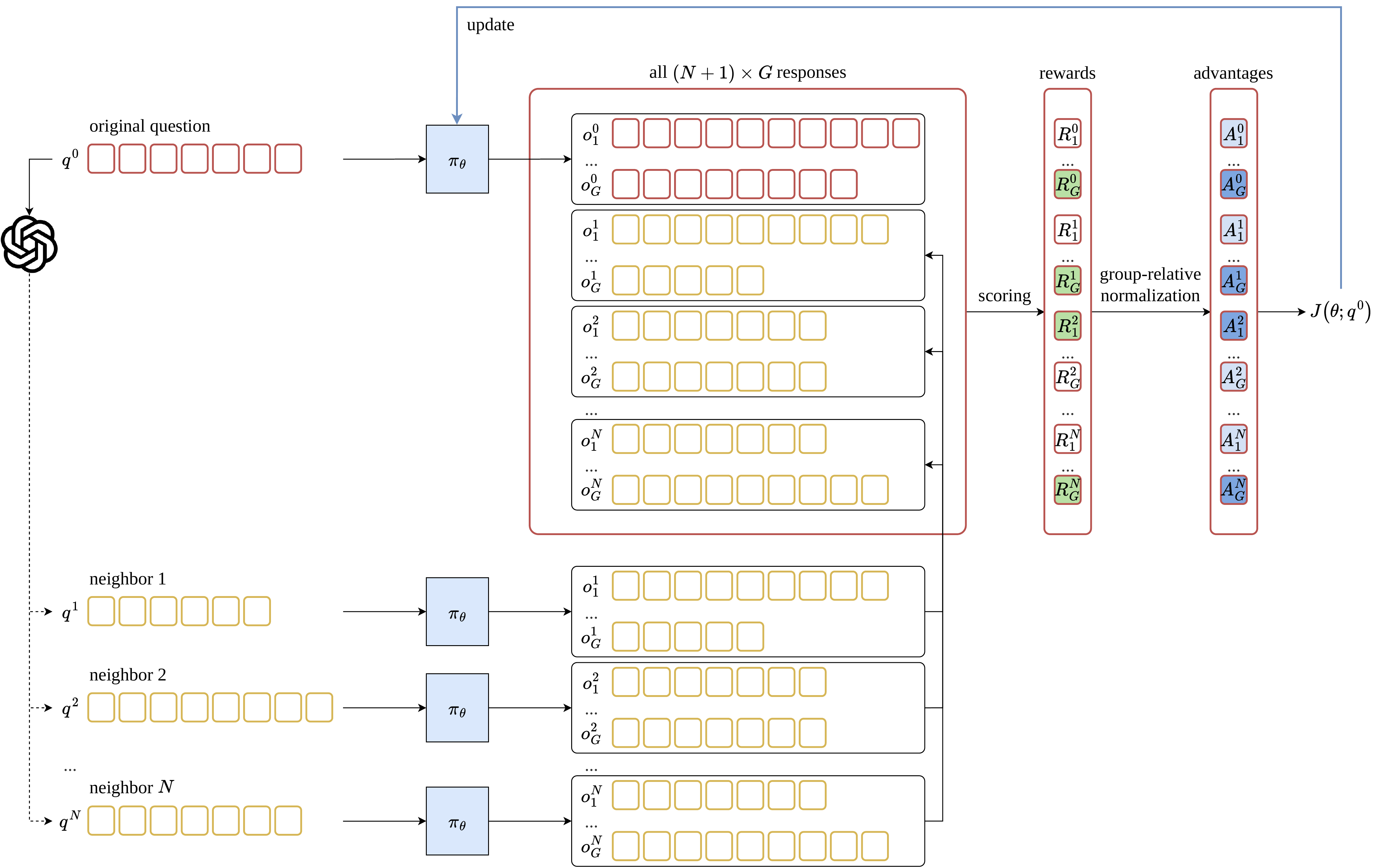}
\caption{
An illustration of our proposed TA-GRPO. Dashed arrows indicate rephrasing the original question. The \textcolor{myblue}{blue} arrow indicates updating the model.
}
\label{fig-2}
\end{figure}

Figure \ref{fig-2} illustrates our proposed Transformation-Augmented GRPO (TA-GRPO) at a glance. First of all, explicitly denote the original question $q$ as $q^0$. We leverage GPT-4-Turbo \citep{achiam2023gpt, hurst2024gpt} and a well-written prompt template with multiple demonstrations (see Appendix \ref{app-a}) to automatically create $N$ problem-equivalent rephrased questions $\{q^n\}^{n=1, ..., N}$ (serving as $N$ “neighbors”). The response group is now formalized as the unified set $\{o_i^n \sim \pi_\theta (\cdot|q^n)\}_{i=1, ..., G}^{n=0, ..., N}$ of size $(N + 1) \times G$. Subsequently, the advantages are computed via \textbf{joint} group-relative reward normalization. Denote all the rewards as $\{R_i^n\}_{i=1, ..., G}^{n=0, ..., N}$, the advantages are:
\begin{equation}
\Big\{A_i^n = \frac{R_i^n - \text{mean}(\{R_i^n\}_{i=1, ..., G}^{n=0, ..., N})}{\text{std}(\{R_i^n\}_{i=1, ..., G}^{n=0, ..., N})+\varepsilon}\Big\}_{i=1, ..., G}^{n=0, ..., N}. 
\label{eqn-3}
\end{equation}
To enable learning from these $(N + 1) \times G$ responses with respect to the original question, TA-GRPO aligns and collects importance sampling ratios under $q^0$ rather than under each $q^n$. This is valid under our problem-equivalence assumption: $\pi_\theta (\cdot|q^n) \approx \pi_\theta (\cdot|q^0)$, which follows from constructing each $q^n$ as a intent-preserving rephrasing of $q^0$. The full TA-GRPO objective is then:
\begin{align}
&\mathfrak{J}(\theta)=-\mathbb{E}_{q^0 \sim \mathcal{D}}[J(\theta; q^0)], \nonumber \\
&J(\theta; q^0) = \left[ \frac{1}{(N + 1) \times G} \sum_{i=1, ..., G}^{n=0, ..., N} \frac{1}{|o_i^n|} \sum_{t=1, ..., |o_i^n|} \Bigg( \text{min}\Big( r_\theta(o_{i, t}^n)A_i^n, \text{clip}\big( r_\theta(o_{i, t}^n), 1-\epsilon, 1+\epsilon \big)A_i^n \Big) \Bigg) \right], \nonumber \\
&\text{where} \quad r_\theta(o_{i, t}^n) = \frac{\pi_\theta(o_{i, t}^n | q^0, o_{i, < t}^n)}{[\pi_\theta(o_{i, t}^n | q^n, o_{i, < t}^n)]_{\text{not\_updated}}}. 
\label{eqn-4}
\end{align}

The design of TA-GRPO directly addresses both gradient vanishing and diversity collapse. Gradient vanishing: By pooling responses across the original question $q^0$ and its $N$ rephrasings, the advantages diminish only when all $(N + 1) \times G$ responses simultaneously succeed or fail, which becomes more unlikely as these rephrasings shift the model’s perceived difficulty. Diversity collapse: In contrast with GRPO, $\{o_i^n\}_{i=1, ..., G}$ are sampled from $\pi_\theta (\cdot|q^n)$ for each $n$; once $\pi_\theta (\cdot|q)$ concentrates and $G$ responses to $q$ are progressively collapsed, other $G$ responses to $q^n$ can still originate from other modes.

\section{Theoretical Analysis}
We first provide a theoretical analysis establishing how TA-GRPO maintains stable model updates. The complete proof is deferred to Appendix \ref{app-b}.

\begin{theorem}[Importance Sampling Unbiasedness \& Variance]
\label{theorem-1}
Let $q^0 \sim \mathcal{D}$ be an original question and $\{q^n\}^{n=1, ..., N}$ be $N$ problem-equivalent rephrased questions. Assume $\sigma$-bounded order-2 Rényi divergence between these rephrasings, i.e., $\max_{n} d_2(\pi_\theta(\cdot|q^0) \parallel \pi_\theta(\cdot|q^n)) \le \sigma$, where $d_2(P \parallel Q) = \log \mathbb{E}_{o \sim Q}[\frac{P(o)^2}{Q(o)^2}]$. The importance sampling ratios $r_\theta(o)$ for responses $o$ drawn from $q^n$ satisfies:
\begin{equation*}
\mathbb{E}_{o \sim \pi_\theta(\cdot|q^n)}[r_\theta(o)] = 1, \quad \text{and} \quad \Var_{o \sim \pi_\theta(\cdot|q^n)}(r_\theta(o)) \le e^{\sigma} - 1.
\end{equation*}
\end{theorem}
Consequently, anchoring importance sampling ratios to $q^0$ produces an unbiased gradient estimate with bounded variance under bounded linguistic perturbations.

Moreover, we establish that question transformation reduces the generalization gap between training and testing. The complete proof is deferred to Appendix \ref{app-c}.
\begin{theorem}[Generalization Gap Bound via Pinsker's Inequality]
\label{theorem-2}
Let $P_{train}(q, t)$ and $P_{test}(q, t)$ denote the training distribution and the testing distribution, respectively, over questions and transformation styles $\mathcal{T} = \{T_0, \dots, T_N\}$. Let $\overline{\rho}_{train}(\theta)$ and $\overline{\rho}_{test}(\theta)$ denote the aggregate success rates under $P_{train}$ and $P_{test}$, respectively. Assume $P_{test} \ll P_{train}$. Then for any $\theta$:
\begin{equation*}
| \overline{\rho}_{test}(\theta) - \overline{\rho}_{train}(\theta) | \le \sqrt{\frac{1}{2} D_{\mathrm{KL}}(P_{test} \parallel P_{train})}.
\end{equation*}
\end{theorem}
Standard GRPO restricts support to $T_0$, making $P_{test} \not\ll P_{train}$ and $D_{\text{KL}}(P_{test} \parallel P_{train}) = \infty$ whenever testing questions appear in different linguistic perturbations. TA-GRPO expands support across $N + 1$ diverse transformation styles, guaranteeing finite \text{KL} divergence and bounding the generalization error.

\section{Experiments}
In this section, we systematically conduct a series of experiments to empirically evaluate our proposed TA-GRPO in enhancing exploration in the reasoning of LLMs. Our evaluation is structured around the following key research questions (RQs):
\begin{itemize}[topsep=0em, leftmargin=*, noitemsep]
\item \textbf{RQ1 - Validity of Question Transformation:} Do the rephrasings \textit{actually} increase the likelihood that the expanded group has responses that receive mixed rewards and encode more diverse reasoning strategies within it?
\item \textbf{RQ2 - Effectiveness:} Does the design of TA-GRPO translate into tangible improvements in challenging reasoning benchmarks over baselines?
\item \textbf{RQ3 - Effectiveness Under the Same Sampling Budget:} Do the improvements simply come from the bigger size of the response group for each training question?
\item \textbf{RQ4 - Data Efficiency:} Can TA-GRPO serve as an alternative to collecting additional high-quality training data, which is more expensive than question transformation? And to what extent?
\end{itemize}
In addition, we provide an ablation study on the design choices of TA-GRPO in Appendix \ref{ablation}.

\begin{figure}[!t]
\centering
\includegraphics[width=\linewidth]{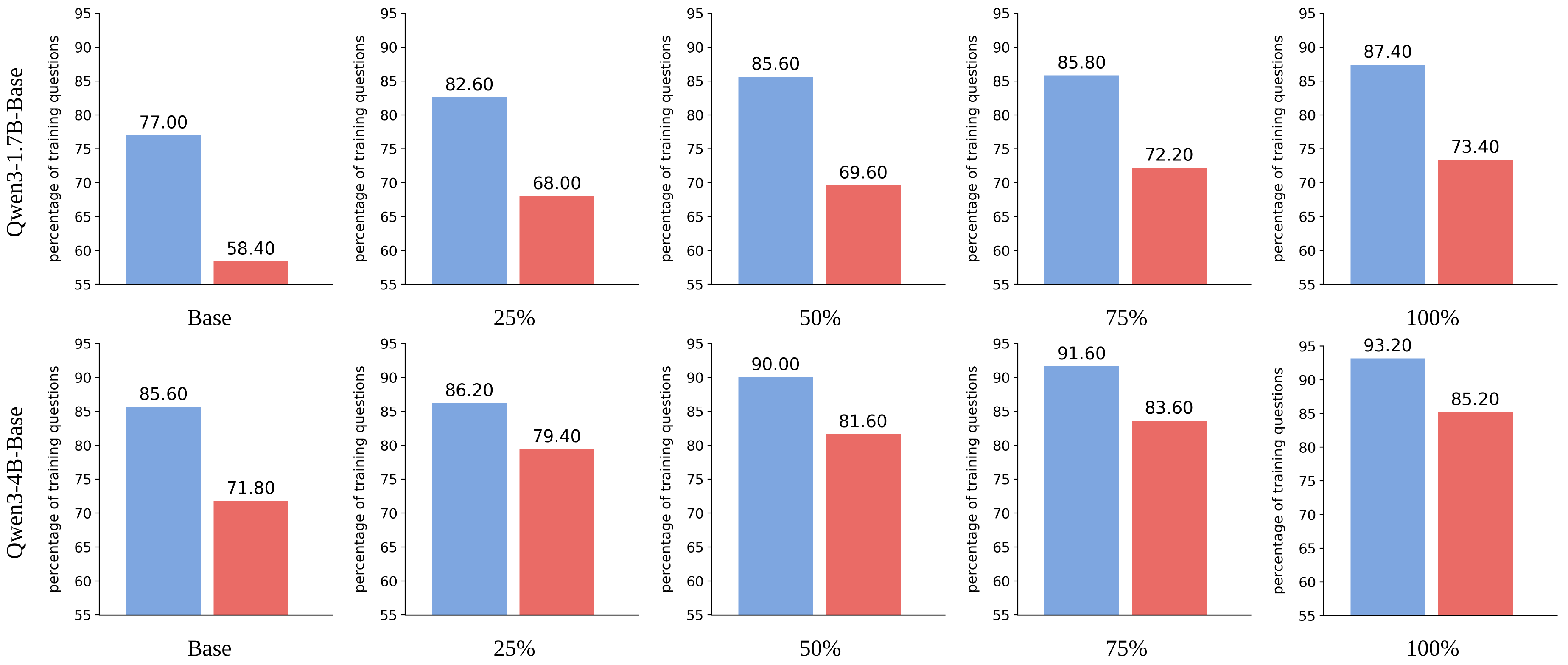}
\caption{
The percentage of training questions for which responses within the group (\colorbox{gblue}{\phantom{a}}) and the expanded group (\colorbox{gred}{\phantom{a}}) receive identical rewards. Base indicates the base models before training, while 25\%, 50\%, 75\%, and 100\% indicate checkpoints throughout the training process.
}
\label{fig-3}
\end{figure}

\begin{figure}[!t]
\centering
\includegraphics[width=\linewidth]{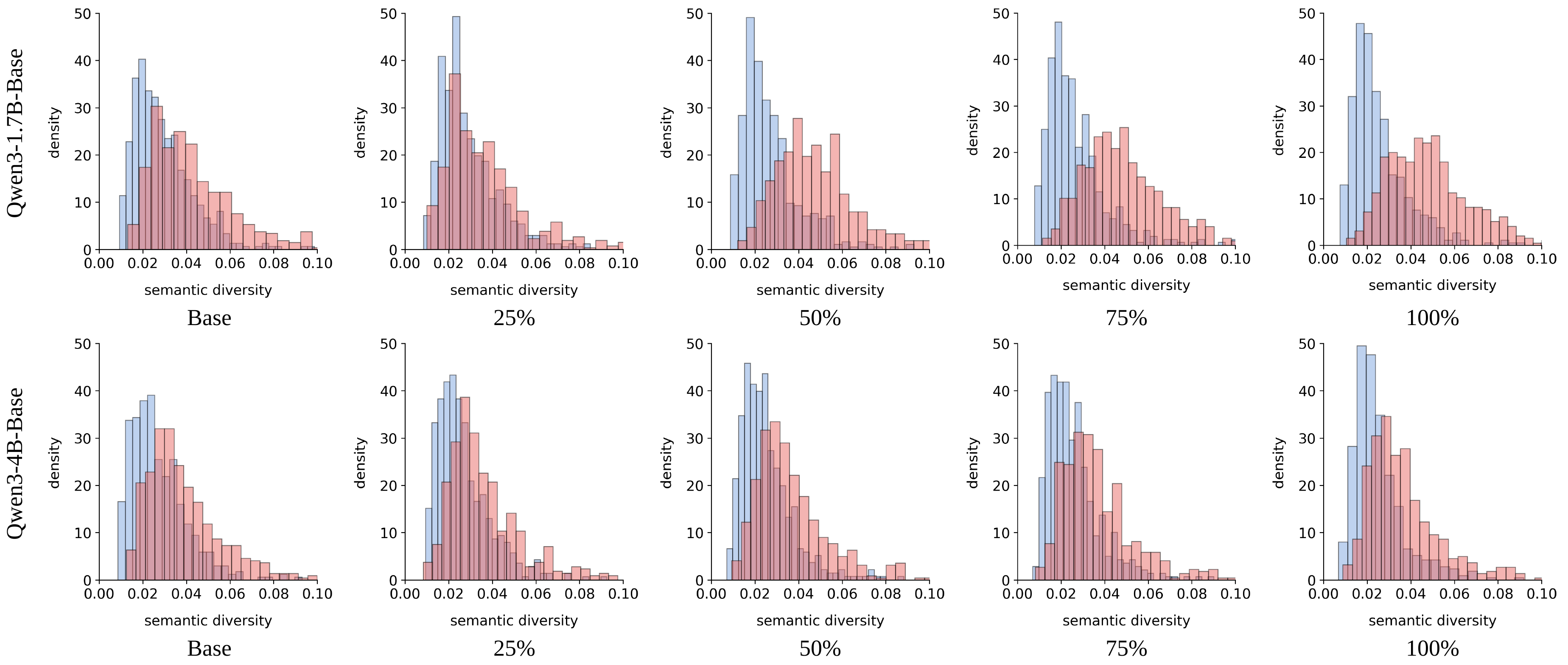}
\caption{
The density distribution of semantic diversity of responses within the group (\colorbox{gblue!50}{\phantom{a}}) and the expanded group (\colorbox{gred!50}{\phantom{a}}). Base indicates the base models before training, while 25\%, 50\%, 75\%, and 100\% indicate checkpoints throughout the training process.
}
\label{fig-4}
\end{figure}

\subsection{Setup}

\textbf{Training Data.} We utilize the popular MATH dataset \citep{hendrycks2021measuring} as training data, which contains 7,498 high school-level mathematics questions.

\textbf{Base LLMs.} We evaluate four base LLMs spanning two model families and a range of scales: Qwen3-1.7B-Base and Qwen3-4B-Base \citep{yang2025qwen3}, and Llama-3.2-1B-Instruct and Llama-3.2-3B-Instruct \citep{grattafiori2024llama}.

\textbf{Baselines.} We compare TA-GRPO against standard GRPO and DAPO \citep{yu2026dapo}, a representative GRPO advance that also targets both gradient vanishing and diversity collapse.

\textbf{Implementation Details.} We implement TA-GRPO and baselines using \texttt{verl} \citep{sheng2025}. Notably, TA-GRPO and baselines consistently let the models generate 8 responses (i.e., $G=8$) to each training question, with a simple instruction prepended to form the input prompt: “\texttt{Please reason step by step, and put your final answer within} $\verb|\|\texttt{boxed}\{\}$”. Other training hyperparameters are provided in Appendix \ref{app-hyperparameters}. Particularly for TA-GRPO, we create 3 rephrasings (i.e., $N=3$) for each training question.

\textbf{Testing Benchmarks.} We evaluate on six benchmarks across two categories: competition-level mathematics benchmarks with AMC \citep{li2024numinamath}, OlympiadBench \citep{he2024olympiadbench}, AIME24, AIME25, and out-of-distribution benchmarks with Minerva \citep{lewkowycz2022}, GPQA-Diamond \citep{rein2024gpqa}.

\textbf{Metric.} We report pass@$k$ \citep{chen2021, yue2025does} with $k$ up to 32, which is sufficiently large to capture the models’ exploration ability. We provide a further discussion of this metric in Appendix \ref{app-metric}.

\subsection{RQ1 - Validity of Question Transformation}
To verify the validity of question transformation, we compare two properties of the response group and the expanded response group, corresponding to before and after applying the transformation, respectively. To assess gradient vanishing, we measure the percentage of training questions for which responses within each group receive identical rewards. To assess diversity collapse, we examine the density distribution of semantic diversity \citep{dang2025weight} of responses within each group, quantified as the average pairwise cosine distance across response embeddings \footnote{The embeddings are extracted by the common sentence embedding model \texttt{sentence-transformers/all-MiniLM-L6-v2}.}.

Figure \ref{fig-3} and Figure \ref{fig-4} present both properties with the base models before training as well as with various checkpoints throughout the training process, using GRPO and TA-GRPO, respectively. For gradient vanishing, Figure \ref{fig-3} shows that, with the base models, responses within the expanded group are far less likely to receive identical rewards (that is, more likely to receive mixed rewards), with the percentage decreasing by 18.6\% on Qwen3-1.7B-Base and 13.8\% on Qwen3-4B-Base, and this gap persists throughout training. For diversity collapse, Figure \ref{fig-4} shows that, with the base models, semantic diversity is markedly greater among responses within the expanded group, as evidenced by the density distribution being more concentrated and skewed to the right, and this gap persists throughout training.

Together, these results confirm that the rephrasings \textit{actually} increase the likelihood that the expanded group has responses that receive mixed rewards and encode more diverse reasoning strategies within it. Importantly, these effects are maintained throughout the training process.

\subsection{RQ2 - Effectiveness}
We next test whether TA-GRPO's design produces improvements in pass@$k$, relative to GRPO and DAPO, evaluated on competition-level mathematics benchmarks and out-of-distribution benchmarks.

\begin{figure}[!t]
\centering\includegraphics[width=\linewidth]{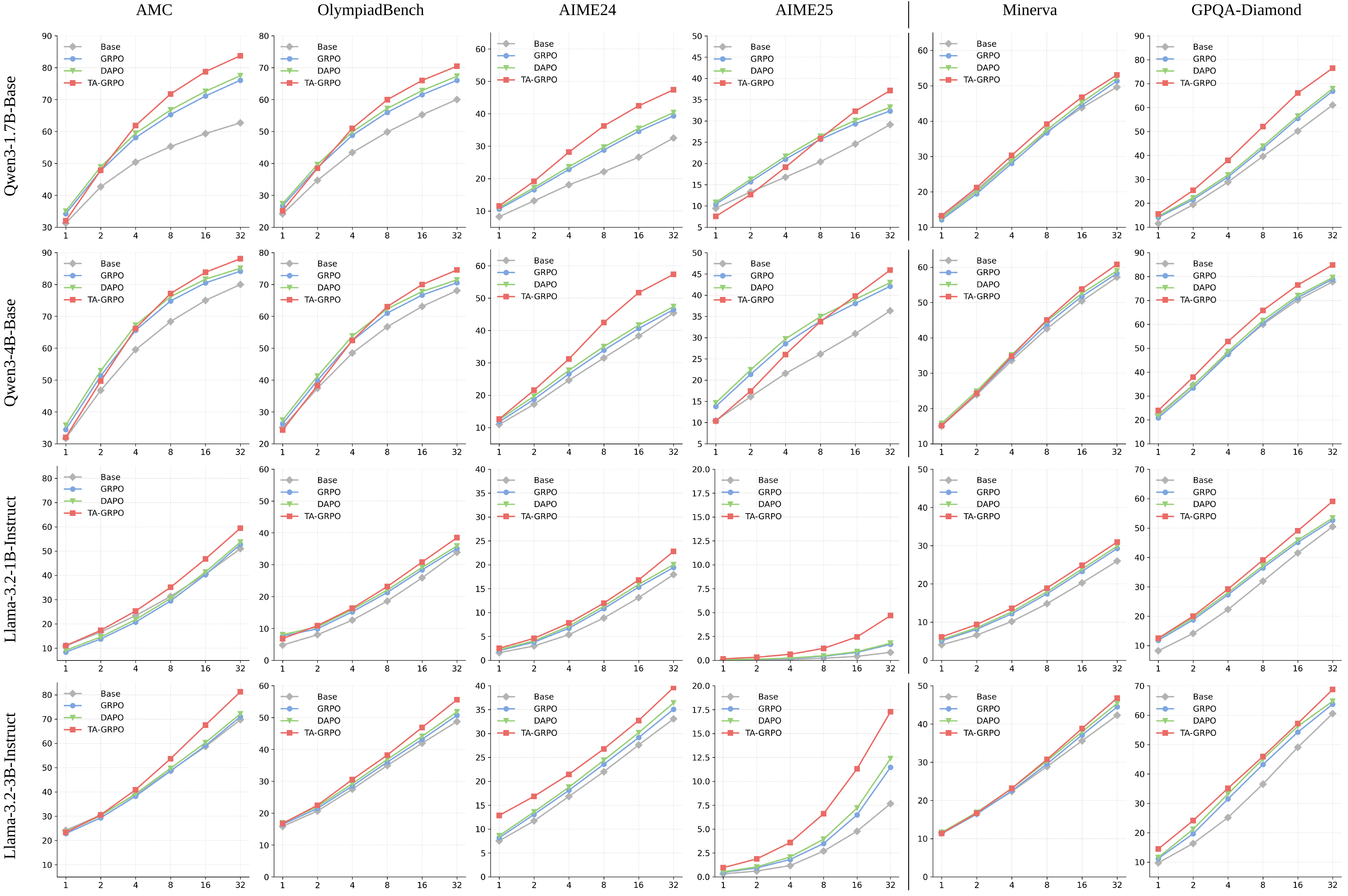}
\caption{The pass@$k$ curves of models on six testing benchmarks. }\label{fig-5}
\end{figure}

Figure \ref{fig-5} displays the pass@$k$ curves of models on six testing benchmarks, and Table \ref{tab-1} reports pass@32 specifically. It can be seen that TA-GRPO consistently improves pass@$k$ across all four models and all six benchmarks, with the gains widening as $k$ grows. The improvements are pronounced on competition-level mathematics benchmarks, where average pass@32 gains over GRPO reach \textbf{6.24}\%/\textbf{5.66}\% on Qwen3-1.7B/4B-Base and \textbf{4.11}\%/\textbf{6.38}\% on Llama-3.2-1B/3B-Instruct. The same trend is seen on out-of-distribution benchmarks, where average pass@32 gains over GRPO reach \textbf{5.68}\%/\textbf{4.23}\% on Qwen3-1.7B/4B-Base and \textbf{4.07}\%/\textbf{3.66}\% on Llama-3.2-1B/3B-Instruct. In contrast, DAPO produces only marginal improvements over GRPO across models and benchmarks, consistent with prior studies that it primarily stabilizes training rather than effectively enhancing exploration in the reasoning of models \citep{yue2025does}.

\begin{table}[!t]
\centering\footnotesize
\caption{
Detailed pass@32 of models on six testing benchmarks. Best results are in \textbf{bold}. 
}
\label{tab-1}
\setlength\tabcolsep{0pt}
\begin{tabular*}{\linewidth}{@{\extracolsep{\fill}} lrccccc@{\hspace{5pt}}|ccc }
\midrule
Model &Method &AMC &OlympiadBench &AIME24 &AIME25 &Average &Minerva &GPQA-Diamond &Average \\
\midrule
\multirow[t]{4}{*}{Qwen3-1.7B-Base}       &Base &62.70          &60.04          &32.54          &29.13          &46.10          &49.64          &61.10          &55.37 \\
&GRPO		                                    &76.06          &66.03          &39.37          &32.34          &53.45          &51.34          &66.91          &59.12 \\
&DAPO		                                    &77.49          &67.31          &40.38          &33.21          &54.60          &52.18          &67.94          &60.06 \\
&TA-GRPO		                                &\textbf{83.74} &\textbf{70.45} &\textbf{47.42} &\textbf{37.14} &\textbf{59.69} &\textbf{53.07} &\textbf{76.52} &\textbf{64.80} \\
\midrule
\multirow[t]{4}{*}{Qwen3-4B-Base}         &Base &79.98          &68.08          &45.44          &36.32          &57.46          &57.19          &77.84          &67.52 \\
&GRPO		                                    &84.16          &70.55          &46.47          &42.11          &60.82          &58.25          &78.96          &68.60 \\
&DAPO		                                    &85.07          &71.46          &47.39          &42.95          &61.72          &58.97          &79.63          &69.30 \\
&TA-GRPO		                                &\textbf{88.08} &\textbf{74.53} &\textbf{57.37} &\textbf{45.92} &\textbf{66.48} &\textbf{60.81} &\textbf{84.85} &\textbf{72.83} \\
\midrule
\multirow[t]{4}{*}{Llama-3.2-1B-Instruct} &Base &50.98          &33.89          &17.96          &00.83          &25.92          &25.98          &50.49          &38.24 \\
&GRPO		                                    &52.81          &35.10          &19.43          &01.67          &27.25          &29.26          &52.68          &40.97 \\
&DAPO		                                    &53.74          &35.88          &20.04          &01.79          &27.86          &29.87          &53.42          &41.64 \\
&TA-GRPO		                                &\textbf{59.46} &\textbf{38.49} &\textbf{22.79} &\textbf{04.68} &\textbf{31.36} &\textbf{30.98} &\textbf{59.11} &\textbf{45.04} \\
\midrule
\multirow[t]{4}{*}{Llama-3.2-3B-Instruct} &Base &69.75          &48.76          &33.07          &07.66          &39.81          &42.30          &60.57          &51.44 \\
&GRPO		                                    &70.99          &50.70          &35.10          &11.47          &42.06          &44.49          &63.75          &54.12 \\
&DAPO                                           &72.18          &51.79          &36.42          &12.38          &43.19          &45.42          &64.78          &55.10 \\
&TA-GRPO	                                    &\textbf{81.25} &\textbf{55.60} &\textbf{39.64} &\textbf{17.28} &\textbf{48.44} &\textbf{46.82} &\textbf{68.74} &\textbf{57.78} \\
\midrule
\end{tabular*}
\end{table}

These results demonstrate that TA-GRPO's design produces consistent and substantial gains in exploration ability across model families and scales, whereas DAPO's benefits remain limited to training stability rather than genuine improvements. Besides that, we observed an interesting finding where, across methods, the pass@$k$ curves exhibit family-dependent shapes: Qwen3-1.7B-Base and Qwen3-4B-Base yield concave curves, whereas Llama-3.2-1B-Instruct and Llama-3.2-3B-Instruct typically yield convex ones. This phenomenon points to underlying behavioral differences between model families that warrant further investigation.

\begin{figure}[!t]
\centering
\includegraphics[width=\linewidth]{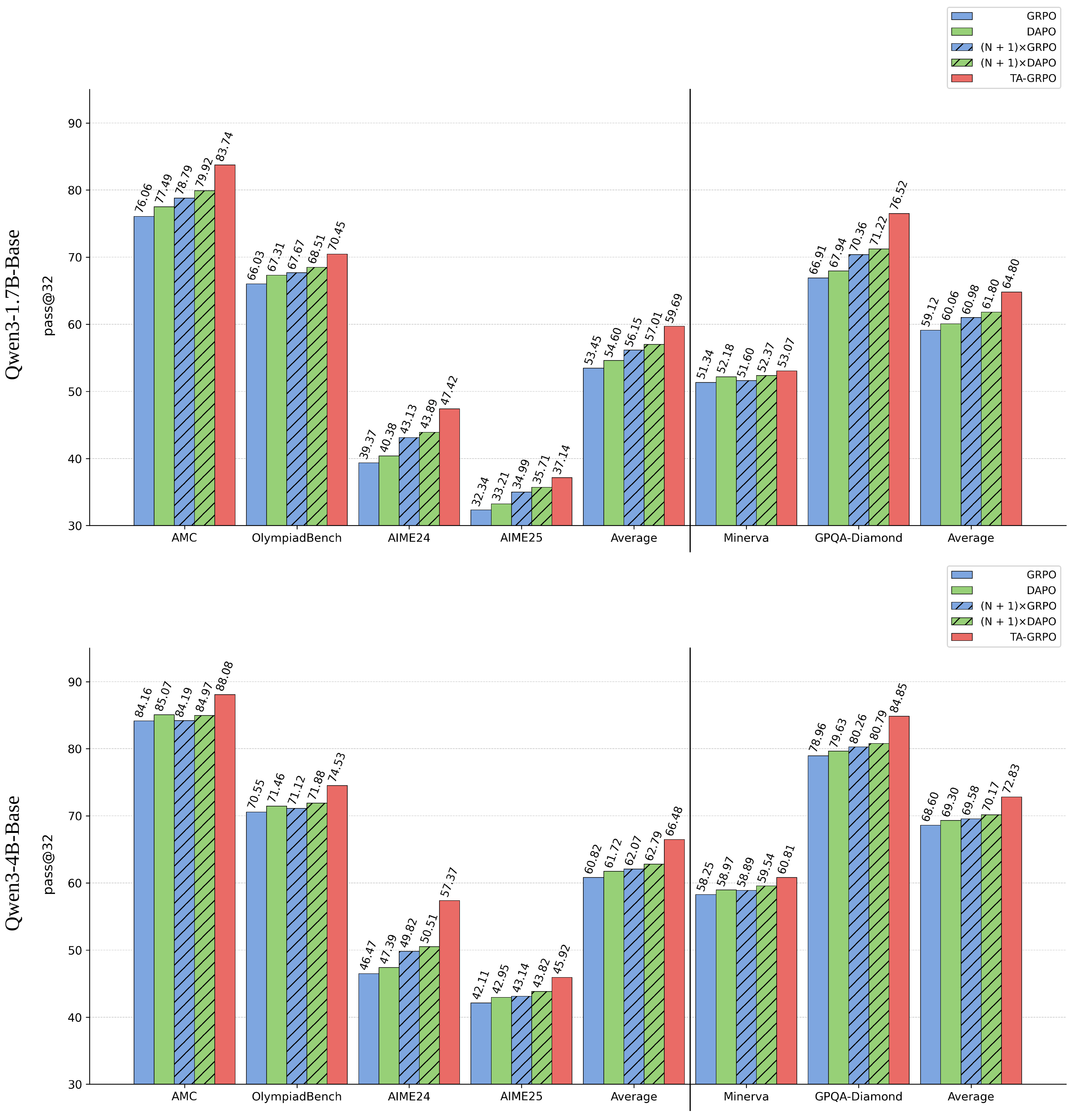}
\caption{Detailed pass@32 of models on six testing benchmarks.}\label{fig-6}
\end{figure}

\subsection{RQ3 - Effectiveness Under the Same Sampling Budget}
One might suspect that such improvements simply stem from learning from a larger response group of size $(N + 1) \times G$ per training question, rather than from TA-GRPO's design itself. To rule out this possibility, we compare TA-GRPO against baselines under a matched sampling budget, denoted as $(N + 1) \times$GRPO and $(N + 1) \times$DAPO. In particular, while TA-GRPO still lets the models generate $G$ responses (i.e., 8) to each training question, $(N + 1) \times$GRPO and $(N + 1) \times$DAPO let the models generate up to $(N + 1) \times G$ responses (i.e., 32), matching the sampling budget that TA-GRPO effectively consumes and ensuring an \underline{apples-to-apples} comparison. All other training hyperparameters are held fixed.

Figure \ref{fig-6} visualizes pass@32 of Qwen3-1.7B-Base and Qwen3-4B-Base on six testing benchmarks. As expected, increasing the response group size per training question elevates both GRPO and DAPO, though only modestly. Importantly, TA-GRPO still maintains clear margins against these upgraded baselines. For instance, on the most challenging AIME24/AIME25 benchmarks, TA-GRPO outperforms upgraded baselines by margins of \textbf{3.53}\%/\textbf{1.43}\% and \textbf{6.86}\%/\textbf{2.10}\% for Qwen3-1.7B-Base and Qwen3-4B-Base, respectively.

With the sampling budget held constant, increasing the response group size per training question would produce modest gains, implying that the remaining gains TA-GRPO delivers indeed stem from its design. In other words, it matters more how responses are sampled and learned from than the quantity.

\subsection{RQ4 - Data Efficiency}
\begin{wrapfigure}{r}{0.48\textwidth}
\centering
\vspace{-32.6pt}
\includegraphics[width=\linewidth]{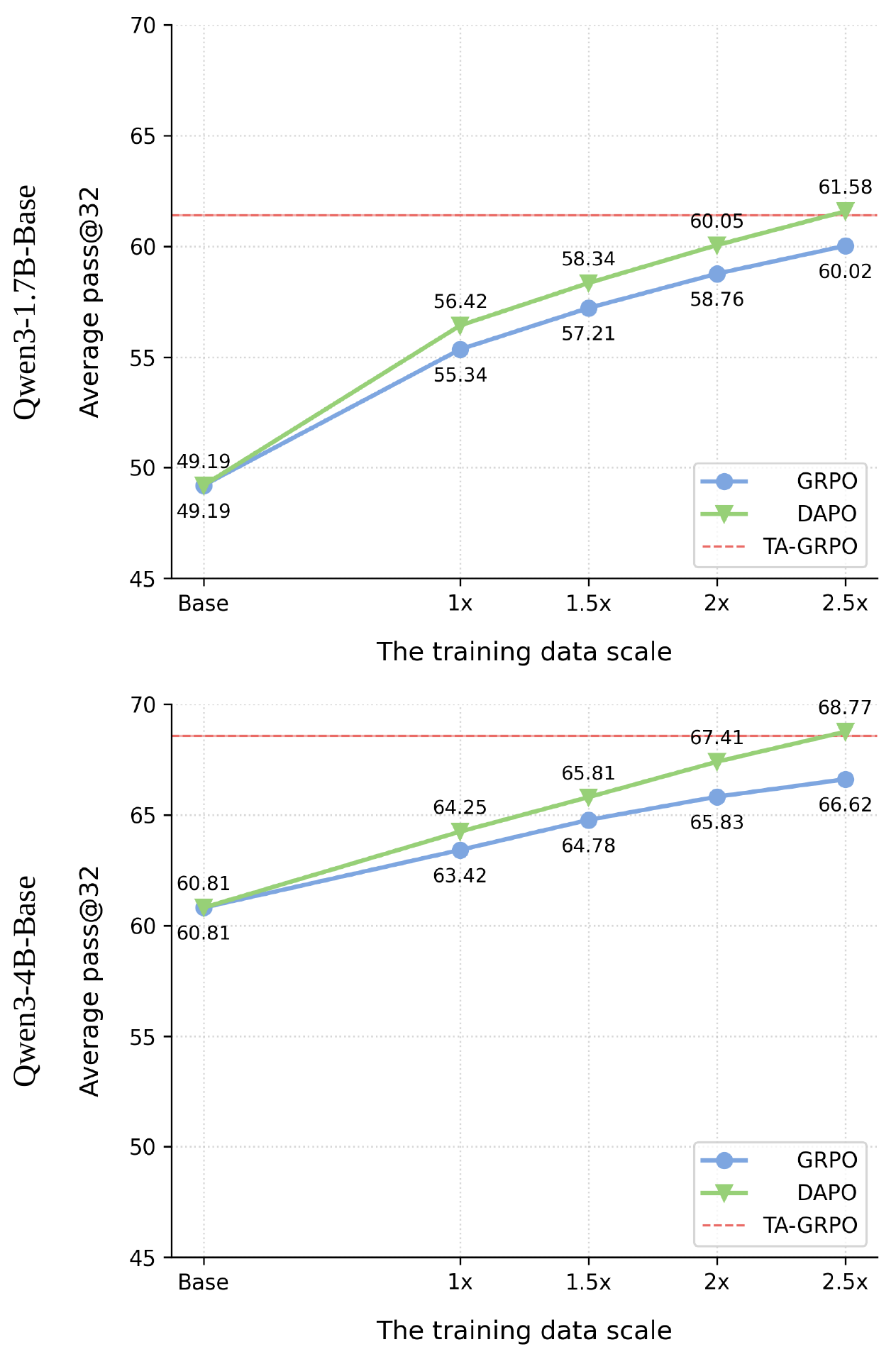}
\caption{Average pass@32 of models across six testing benchmarks.}\label{fig-7}
\end{wrapfigure}
Lastly, we investigate whether TA-GRPO, with multiple rephrasings for each training question, can serve as a substitute for collecting additional high-quality training data, a process that is typically more expensive and time-consuming. To address this RQ, we draw on the large-volume OpenR1-MATH-220k dataset \citep{openr1} (220k mathematics questions collected from diverse sources), randomly selecting and incrementally appending chunks of 3,749 questions (half the size of the MATH dataset) to the training data to simulate training data scales of 1×, 1.5×, 2×, and so forth. We then compare TA-GRPO trained with the MATH dataset against baselines trained with this more data.

Figure \ref{fig-7} visualizes average pass@32 of Qwen3-1.7B-Base and Qwen3-4B-Base across six testing benchmarks. For both GRPO and DAPO, increasing training data returns roughly linear gains in pass@32, as expected. Notably, DAPO becomes increasingly useful with more questions from the OpenR1-MATH-220k dataset. Remarkably, TA-GRPO trained with the MATH dataset (i.e., 1× data) matches the pass@32 of baselines trained with up to 2.5× more data, reaching an average of \textbf{61.39}\% and \textbf{68.59}\% for Qwen3-1.7B-Base and Qwen3-4B-Base, respectively.

These results suggest that TA-GRPO can practically substitute for roughly 1.5× additional collected training data, offering a much cheaper and faster alternative to scaling up data collection while achieving comparable gains in pass@32.

\section{Limitations}
This section discusses the limitations of our proposed TA-GRPO. First, the method relies on GPT-4-Turbo to create $N$ problem-equivalent rephrased questions for each training question; while this ensures rephrasing quality, it incurs a modest API cost. Replacing GPT-4-Turbo with a comparable open-weight model would be beneficial and should be investigated. Second, because TA-GRPO leverages these rephrasings, it introduces extra training-time overhead relative to the GRPO and DAPO baselines (see Appendix \ref{app-hyperparameters}). However, this overhead is offset by consistent and substantial gains in exploration ability across model families and scales, with TA-GRPO outperforming even the matched-budget baselines $(N + 1) \times$GRPO and $(N + 1) \times$DAPO, which incur comparable training time (apples-to-apples comparison).

\section{Conclusion}
We introduced TA-GRPO, a simple yet effective extension of GRPO that mitigates gradient vanishing and diversity collapse via question transformation, pooling responses across each training question together with its $N$ rephrasings into an expanded group, jointly computing advantages and anchoring importance sampling ratios to the original question. Across four LLMs, TA-GRPO consistently improves pass@$k$ on all six benchmarks, and matches the pass@32 of baselines trained with up to 2.5× more data. TA-GRPO highlights the question and response levels as additional axes for improving GRPO, complementing the objective axis explored by prior work, and is thus orthogonal to and compatible with existing GRPO advances.

\bibliography{main}
\bibliographystyle{tmlr}

\clearpage
\appendix
\section{Prompt Template}
\label{app-a}

\begin{tcolorbox}[left=5pt, right=5pt, top=5pt, bottom=5pt, fonttitle=\normalsize, colback=red!5, colframe=red!50, title={
Box 1: The prompt template with multiple demonstrations for rephrasing a given \texttt{\underline{question}} using GPT-4-Turbo.
}]
\begin{minipage}{\textwidth}
\normalsize
You are a helpful AI assistant. Given a question, your task is to rephrase it from different perspectives while preserving the underlying intent. You should apply linguistic perturbations (changes in wording, format, or information order) rather than plain changes like variable renaming. Follow the given examples:
\\\\
Question: Kevin Kangaroo begins hopping on a number line at 0. He wants to get to 1, but he can hop only $\frac{1}{3}$ of the distance. Each hop tires him out so that he continues to hop $\frac{1}{3}$ of the remaining distance. How far has he hopped after five hops? Express your answer as a common fraction.
\\Rephrased: Starting at 0 on a number line, Kevin Kangaroo aims to reach 1 by hopping only 1/3 of the distance each time. As he gets tired, he continues to hop 1/3 of the remaining distance. After completing five hops, what is the total distance he has covered, expressed as a common fraction?
\\\\
Question: What is the area of the region defined by the equation $x^2+y^2 - 7 = 4y-14x+3$?
\\Rephrased: Determine the area of the region described by the equation $x^2+y^2 - 7 = 4y-14x+3$?
\\\\
Question: If $x^2+y^2=1$, what is the largest possible value of $|x|+|y|$?
\\Rephrased: What is the maximum value possible for $|x| + |y|$ if $x^2 + y^2 = 1$?
\\\\
Question: If $f(x)=\frac{ax+b}{cx+d}, abcd\not=0$ and $f(f(x))=x$ for all $x$ in the domain of $f$, what is the value of $a+d$?
\\Rephrased: Given that $f(x) = \frac{ax + b}{cx + d}$, with all variables not equal to 0, and $f(f(x)) = x$ for all x within the domain of f, what is the value of $a + d$?
\\\\
Question: A math teacher requires Noelle to do one homework assignment for each of the first five homework points she wants to earn; for each of the next five homework points, she needs to do two homework assignments; and so on, so that to earn the $n^{\text{th}}$ homework point, she has to do $n\div5$ (rounded up) homework assignments. For example, when she has 11 points, it will take $12\div5=2.4\rightarrow3$ homework assignments to earn her $12^{\text{th}}$ point. What is the smallest number of homework assignments necessary to earn a total of 25 homework points?
\\Rephrased: Noelle's math teacher has a system where she needs to complete one homework assignment for each of the first five points, two assignments for each of the next five points, and so on, with the number of assignments required for the nth point being n divided by 5 (rounded up). For instance, to get her 12th point, she needs to complete 3 assignments (12/5 = 2.4, rounded up to 3). What is the minimum number of homework assignments Noelle must complete to earn a total of 25 points?
\\\\
Question: The quadratic equation $x^2+mx+n=0$ has roots that are twice those of $x^2+px+m=0$, and none of $m$, $n$, and $p$ is zero. What is the value of $n/p$?
\\Rephrased: For the quadratic equation $x^2 + mx + n = 0$, the roots are twice the roots of $x^2 + px + m = 0$. None of the variables $m$, $n$, and $p$ are zero. What is the value of $n/p$?
\\\\
Question: Expand $(2z^2 + 5z - 6)(3z^3 - 2z + 1)$.
\\Rephrased: What is the expanded form of $(2z^2 + 5z - 6)(3z^3 - 2z + 1)$?
\\\\
Question: Find the mean of all solutions for $x$ when $x^3 + 3x^2 - 10x = 0$.
\\Rephrased: What is the average of all the solutions for $x$ in the equation $x^3 + 3x^2 - 10x = 0$?
\\\\
Question: \{\texttt{\underline{question}}\}
\\Rephrased: 
\end{minipage}
\end{tcolorbox}

\section{Proof of Theorem \ref{theorem-1}}
\label{app-b}
\begin{proof}
\phantom{}

For every $o \in \mathcal{O}$, we have $\pi_\theta(o|q^0) > 0$ and $\pi_\theta(o|q^n) > 0$, since both are softmax distributions with full support. Hence the importance sampling ratios $r_\theta(o) = \frac{\pi_\theta(o|q^0)}{\pi_\theta(o|q^n)}$ are well-defined, finite, and strictly positive on all of $\mathcal{O}$. By definition of $\pi_\theta(o|q^n)$:
\begin{align}
\mathbb{E}_{o \sim \pi_\theta(\cdot|q^n)}[r_\theta(o)] &= \mathbb{E}_{o \sim \pi_\theta(\cdot|q^n)} \left[ \frac{\pi_\theta(o|q^0)}{\pi_\theta(o|q^n)} \right] \nonumber \\
&= \sum_o \pi_\theta(o|q^n) \frac{\pi_\theta(o|q^0)}{\pi_\theta(o|q^n)} \nonumber \\
&= \sum_o \pi_\theta(o|q^0) \nonumber \\
&= 1, \nonumber 
\end{align}
and
\begin{align}
\Var_{o \sim \pi_\theta(\cdot|q^n)}(r_\theta(o)) &= \mathbb{E}_{o \sim \pi_\theta(\cdot|q^n)}[r_\theta(o)^2] - (\mathbb{E}_{o \sim \pi_\theta(\cdot|q^n)}[r_\theta(o)])^2 \nonumber \\
&= \mathbb{E}_{o \sim \pi_\theta(\cdot|q^n)} \left[ \frac{\pi_\theta(o|q^0)^2}{\pi_\theta(o|q^n)^2} \right] - 1 \nonumber \\
&= \exp\big(d_2(\pi_\theta(\cdot|q^0) \parallel \pi_\theta(\cdot|q^n))\big) - 1, \nonumber 
\end{align}
where $d_2$ is the Rényi divergence of order 2. Under $\sigma$-bounded order-2 Rényi divergence, $\exp(d_2) \le e^\sigma$, yielding $\Var_{o \sim \pi_\theta(\cdot|q^n)}(r_\theta(o)) \le e^\sigma - 1$.

\end{proof}

\section{Proof of Theorem \ref{theorem-2}}
\label{app-c}
\begin{proof}
\phantom{}

With a binary reward function $\mathcal{R}: \mathcal{Q} \times \mathcal{O} \to \{0, 1\}$ over questions $q \in \mathcal{Q}$ and sampled responses $o \in \mathcal{O}$, the expected success rate $g_\theta(q, t) \in [0, 1]$ for any pair $(q, t) \in \mathcal{Q} \times \mathcal{T}$ is defined as:
$$g_\theta(q, t) := \mathbb{E}_{o \sim \pi_\theta(\cdot \mid t(q))}[\mathcal{R}(q, o)]$$
The performance difference (expected generalization gap) between training and testing is:
$$\overline{\rho}_{test}(\theta) - \overline{\rho}_{train}(\theta) = \int_{\mathcal{Q} \times \mathcal{T}} g_{\theta}(q, t) \, d(P_{test} - P_{train})$$
Since $0 \le g_{\theta}(q, t) \le 1$, by the definition of Total Variation distance:
$$D_{\text{TV}}(P_{test}, P_{train}) = \sup_{g: \mathcal{X} \to [0, 1]} \left | \int g \, d(P_{test} - P_{train}) \right |$$
Substituting this into the equality gives:
$$| \overline{\rho}_{test}(\theta) - \overline{\rho}_{train}(\theta) | \le D_{\text{TV}}(P_{test}, P_{train})$$
Applying Pinsker's Inequality yields:
$$| \overline{\rho}_{test}(\theta) - \overline{\rho}_{train}(\theta) | \le \sqrt{\frac{1}{2} D_{\mathrm{KL}}(P_{test} \parallel P_{train})}$$

\end{proof}

\section{Training Hyperparameters}
\label{app-hyperparameters}
We implement TA-GRPO and baselines using \texttt{verl} \citep{sheng2025}. Key training hyperparameters for reproducibility are listed in Table \ref{tab-2}. We run all experiments on 2 NVIDIA A100 80GB GPUs, and report the corresponding training times in Table \ref{tab-3}. Note that we use the MATH dataset \citep{hendrycks2021measuring} as training data, which contains 7,498 high school-level mathematics questions that typically do not require lengthy responses. Empirically, our training logs indicate that the response truncation rate was only around \underline{\textcolor{red}{1\%}} when capping the maximum response length at 4096 tokens. This indicates that a max response length of 4096 does not affect our observations and conclusions while reducing computational expense.

\begin{table}[!ht]
\centering\small
\caption{Key Training Hyperparameters.}
\label{tab-2}
\begin{tabular}{lr}
\midrule
max prompt length           & 512   \\
max response length         & 4096  \\
temperature                 & 1.0   \\
learning rate               & 1e-6  \\
KL loss coefficient         & 0.0   \\
clip ratio (lower)          & 0.2   \\
clip ratio (upper)          & 0.2   \\
batch size                  & 128   \\
mini batch size             & 64    \\
micro batch size per GPU    & 8     \\
total epochs                & 2     \\
\midrule
\end{tabular}
\end{table}
\begin{table}[!ht]
\centering\small
\caption{Training Times of TA-GRPO and Baselines.}
\label{tab-3}
\begin{tabular}{lrr}
\midrule
Model &Method &Training Times \\
\midrule
\multirow[t]{5}{*}{Qwen3-1.7B-Base} &GRPO & $\sim$6  hours \\
&DAPO                                     & $\sim$6  hours \\
&$(N + 1) \times$GRPO                     & $\sim$24 hours \\
&$(N + 1) \times$DAPO                     & $\sim$24 hours \\
&TA-GRPO                                  & $\sim$24 hours \\
\midrule
\multirow[t]{5}{*}{Qwen3-4B-Base}   &GRPO & $\sim$12 hours \\
&DAPO                                     & $\sim$12 hours \\
&$(N + 1) \times$GRPO                     & $\sim$48 hours \\
&$(N + 1) \times$DAPO                     & $\sim$48 hours \\
&TA-GRPO                                  & $\sim$48 hours \\
\midrule
\end{tabular}
\end{table}

\section{On the pass@\texorpdfstring{$k$}{k} Metric}
\label{app-metric}
Here, we provide a thorough discussion of the pass@$k$ metric used in the paper.

\textbf{Definition.} pass@$k$ \citep{chen2021} measures the probability that at least one of $k$ independently sampled responses from the model succeeds (i.e., produces a correct solution) for a given question. Formally, given the question and $n$ responses to it, of which $c$ succeed, an unbiased estimator of pass@$k$ is:
$$\text{pass@}k = \mathbb{E}_{\text{questions}}\left[ 1 - \frac{\binom{n-c}{k}}{\binom{n}{k}} \right].$$
When $k = 1$, the metric reduces to the average success rate under temperature sampling; when $k > 1$, it captures the model's ability to produce a correct solution at least once across multiple attempts. We compute the metric using a temperature of 0.6 and a top-$p$ of 0.95.

\textbf{Connection to the Exploitation–Exploration Trade-off.} The choice of $k$ encodes a fundamental trade-off in how we evaluate the model:
\begin{itemize}[topsep=0em, leftmargin=*, noitemsep]
\item pass@1 rewards \textit{Exploitation}: The model must concentrate its sampling distribution mass on the correct solution, and the evaluation reflects what a user would experience from a single query.
\item pass@$k$ for a larger $k$ rewards \textit{Exploration}: The model is credited for any correct solution within its sampling distribution, even if that correct solution is rare.
\end{itemize}

\textbf{Misleading When $k$ Is Too Large.} However, an excessively large $k$ inflates pass@$k$ by giving flawed reasoning more chances to stumble onto a correct solution while also amplifying noise and weakening the metric's ability to discriminate between models. Moderate values of $k$, typically between 2 and 32, keep the metric focused on reasoning capabilities rather than repeated guessing and better preserve differences between models.

Therefore, we use pass@$k$ and limit our evaluation to $k = 32$.

\section{Ablation Study}
\label{ablation}
\begin{wrapfigure}{r}{0.48\textwidth}
\centering
\vspace{-32.6pt}
\includegraphics[width=\linewidth]{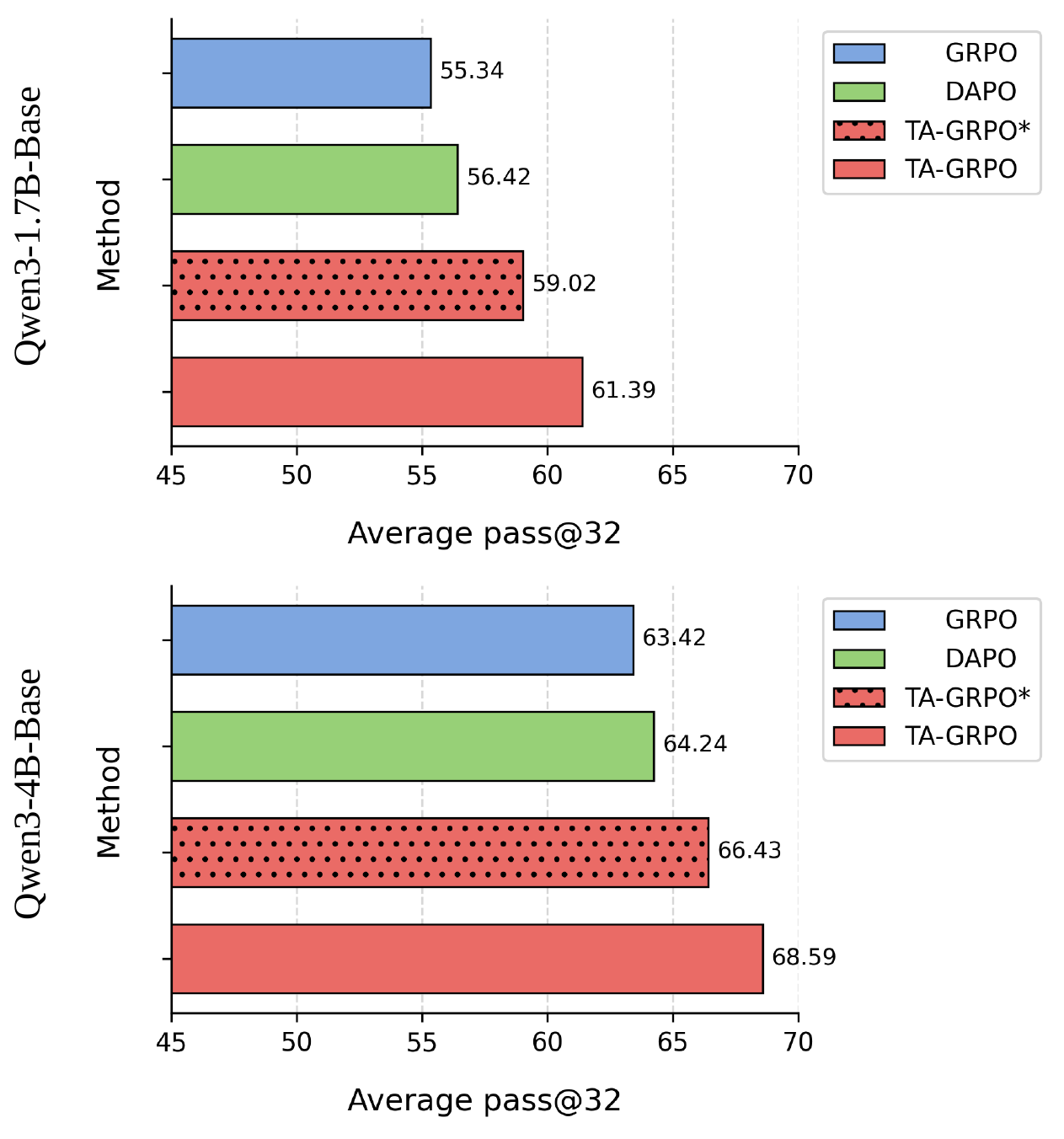}
\caption{
Average pass@32 of models across six testing benchmarks.
}
\label{fig-8}
\end{wrapfigure}
TA-GRPO comprises two design choices: question transformation and algorithmic modifications to standard GRPO (pooling responses across each training question and its $N$ rephrasings into an expanded group, jointly computing advantages and anchoring importance sampling ratios to the original question). To isolate the contribution of the algorithmic modifications, we remove them and denote the resulting simplified version as TA-GRPO*. All other training hyperparameters are held fixed.

Figure \ref{fig-8} visualizes average pass@32 of Qwen3-1.7B-Base and Qwen3-4B-Base across six testing benchmarks. TA-GRPO achieves gains of \textbf{4.97}\% and \textbf{4.35}\% over the baselines on Qwen3-1.7B-Base and Qwen3-4B-Base, respectively, of which the algorithmic modifications alone contribute \textbf{2.37}\% and \textbf{2.16}\% — roughly half of the total gain in each case.

These results indicate that question transformation and the algorithmic modifications contribute complementary, roughly equal shares to TA-GRPO's overall gains, confirming that both design choices are essential to its effectiveness.

\end{document}